%% file: main.tex
\documentclass[letterpaper, 10 pt, conference]{ieeeconf}  

\IEEEoverridecommandlockouts                              

\usepackage{amsmath,amssymb}
\usepackage{bm}

\usepackage[noadjust]{cite}
\usepackage{todonotes}
\usepackage{makecell}
\usepackage{siunitx}
\usepackage{booktabs} 
\usepackage{multirow} 
\usepackage{algorithm}
\usepackage{algpseudocode}
\usepackage{graphicx}
\usepackage[hidelinks]{hyperref}
\usepackage{caption}
\usepackage{subcaption}
\usepackage{cleveref}
\newtheorem{definition}{Definition}

\usepackage{stackengine}
\crefname{section}{Section}{Sections}
\crefname{subsection}{Section}{Sections}
\crefname{definition}{Definition}{Definitions}
\crefname{proposition}{Proposition}{Propositions}
\crefname{lemma}{Lemma}{Lemmas}
\crefname{theorem}{Theorem}{Theorems}
\crefname{corollary}{Corollary}{Corollaries}
\crefname{example}{Example}{Examples}
\crefname{figure}{Figure}{Figures}
\crefname{assumption}{Assumption}{Assumptions}
\crefname{remark}{Remark}{Remarks}
\crefname{running}{Running Example}{Running Examples}
\crefname{algorithm}{Algorithm}{Algorithms}

\title{\LARGE \bf
Stealthy Perception-based Attacks on Unmanned Aerial Vehicles*
}

\author{Amir Khazraei, Haocheng Meng, and Miroslav Pajic 
\thanks{*This work is sponsored in part by the ONR under the agreementS N00014-20-1-2745 and N00014-23-1-2206, AFOSR award number FA9550-19-1-0169, NSF CNS-1652544 award as well as the National AI Institute for Edge Computing Leveraging Next Generation Wireless Networks, Grant CNS-2112562.}
\thanks{The authors are with the Department of Electrical and Computer
Engineering, Duke University, Durham, NC 27708 USA (e-mail:
amir.khazraei@duke.edu, haocheng.meng@duke.edu, miroslav.pajic@duke.edu).}
}

\begin{document}

\maketitle
\thispagestyle{empty}
\pagestyle{empty}

\begin{abstract}

In this work, we study vulnerability of 
unmanned aerial vehicles (UAVs) to stealthy  
attacks on perception-based control. To guide our analysis, we consider two specific missions: ($i$) ground vehicle tracking (GVT), and ($ii$) vertical take-off and landing (VTOL) of a quadcopter on a moving ground vehicle. Specifically, we introduce a method to consistently attack both the sensors measurements and camera images over time, in order to cause control performance degradation 
(e.g., by failing the mission) while remaining stealthy (i.e., undetected by the 
deployed 
anomaly detector).  Unlike 
existing attacks that mainly rely on vulnerability of deep neural networks to small input perturbations (e.g., by adding small patches and/or noise to the images), we show that stealthy yet effective attacks can be designed by changing 
images  of the ground vehicle's landing markers as well as suitably falsifying sensing data. We illustrate the effectiveness of our attacks in Gazebo 3D robotics simulator.
%
\end{abstract}





\input{Introduction}

\input{Related_work}
\input{framework_procedure}

\input{Evaluation}

\input{Conclusion}

\bibliographystyle{IEEEtranMod}




\end{document}

%% file: Introduction.tex
\vspace{-6pt}
\section{Introduction}
\label{sec:introduction}
\vspace{-2pt}

Recent years have witnessed significant research attention focused on unmanned aerial vehicles (UAVs) in a variety of military and civilian applications 
\cite{beloev2016review}. Also, recent progress in computer vision has resulted in a new generation of controllers that utilize visual data for decision making and control. Popular control tasks that incorporate visual data into the UAV's control-loop are vertical take-off and landing (VTOL) and ground vehicle tracking (GVT). These tasks have been widespread in the areas of rescue and reconnaissance, package delivery, inspection in 
hazardous environments, etc. Yet, despite their wide application, vulnerability of such systems to cyber attacks has not been well studied.  

Quadcopter UAV's have been shown vulnerable to adversarial attacks, illustrating security 
challenges in these systems (e.g.,~\cite{rodday2016exploring,javaid2012cyber,hartmann2013vulnerability}). For example,~\cite{rodday2016exploring} has shown that an attacker could perform a Man-in-the-Middle attack, and inject false data and control commands to 
the compromised UAV, 
with catastrophic impact on the system and even human life. 
Thus, in this work, our goal is to evaluate the physical impact  of these attacks (i.e., on the UAV's behavior), focusing on the false data-injection attacks that compromise both camera images and sensor measurements used for control. 

Starting from~\cite{szegedy2013intriguing}, most 
existing adversarial attacks on images leverage vulnerability of deep neural networks to small changes in their input. The idea is to add imperceptible noise to the original image, causing the deep neural network's output to deviate from its original output~\cite{patel2021overriding}.  
This idea has been applied to UAVs in tracking~\cite{fu2022ad2attack} and control tasks~\cite{tian2021adversarial,patel2021overriding}. For example, \cite{fu2022ad2attack} proposes adversarial attacks that add noise to the images to cause tracking drift in the bounding box around the target objects; in~\cite{tian2021adversarial}, noise is added to the image to change the steering angle and collision probabilities, degrading the UAV navigation performance.

Yet, despite their success in disrupting the controller performance, these attacks are not designed to be stealthy. Most existing UAVs are equipped with an anomaly detector (e.g., $\chi^2$ detector based on the extended Kalman filter) that use the raw sensor measurements (and possibly the output of a vision module like position/pose estimation) to detect the presence of abnormal behaviors; thus, effectively limiting impact of the attacks.  
For non-perception control systems,  stealthy attacks 
have been well-defined in e.g.,~\cite{mo2009secure,mo2010false,teixeira2012revealing,khazraei2022attack,smith2015covert,bai2017data,sui2020vulnerability,khazraei2022resiliency,khazraei2020perfect,pajic_tcns17,jovanov_tac19,khazraei2022learning}, including replay~\cite{mo2009secure}, covert~\cite{smith2015covert}, zero-dynamic~\cite{teixeira2012revealing}, and false data injection attacks~\cite{mo2010false,jovanov_tac19, khazraei2022attack}. 
However, in addition to not covering perception, these works only focus on LTI systems and controllers, limiting their use on UAVs. 

On the other hand, control performance degradation caused by an attack on camera images used for control will be reflected on the employed physical sensor measurements (e.g., GPS and IMU). Hence, as we will show in the paper, if the sensor data is not suitably falsified, attacks on camera images will likely be detected. 
Consequently, it is unclear how vulnerable are the UAV controllers that employ both visual and `traditional' sensing data, and whether it is possible to launch stealthy and effective attacks that significantly degrade control performance while staying undetected.  

To answer this question, in this work, we introduce a general framework to design false data injection attacks that consistently falsify both camera images and physical sensors data, causing significant deviations 
from the trajectory of the non-compromised drone, while being stealthy from the system's intrusion detector. Specifically, we investigate attacks on two different mission tasks -- VTOL and GVT, where a vision-based controller is used to navigate the drone. For both tasks, we consider a squared fiducial marker ArUco~\cite{garrido2014automatic} on a moving vehicle, with a correlation filter being used to detect the marker and estimate the relative position and the heading of the drone with respect to the marker. 

To show stealthiness of our attacks, in our experiments, we consider two widely-used anomaly detectors -- $\chi^2$ 
and deep learning-based detectors. We show that instead of adding noise to the image as commonly done today for non-control applications,  {the attacker should manipulate the scene geometry of the current image in a way that is governed by the UAVs dynamics}. For example, for a moving vehicle tracking, our 
attack strategy 
starts with a desired false relative position. 
Then, the attacker constructs 
a false image that renders the falsified relative position of the marker. The desired false relative position is not chosen arbitrarily; instead, it is governed by the system dynamics to be stealthy from any intrusion detector. We will show that while having a physical impact on the drone (deviating the drone from its desired trajectory in GVT and landing in a wrong place in VTOL), the resulting attacks are stealthy from any considered anomaly detector. 

This paper is organized as follows. Section~\ref{sec:model} introduces the system model, before we present a formal attack model, capturing the required attack impact and stealthiness constraints, as well as a procedure to design such attacks (Section~\ref{sec:attacks}).  Section~\ref{sec:result} demonstrates effectiveness of the develop attacks, before  concluding remarks in Section~\ref{sec:conclusion}.

%% file: framework_procedure.tex
\section{System Model}
\label{sec:model}

\begin{figure}[!t]
\centerline{\includegraphics[width=0.92\columnwidth]{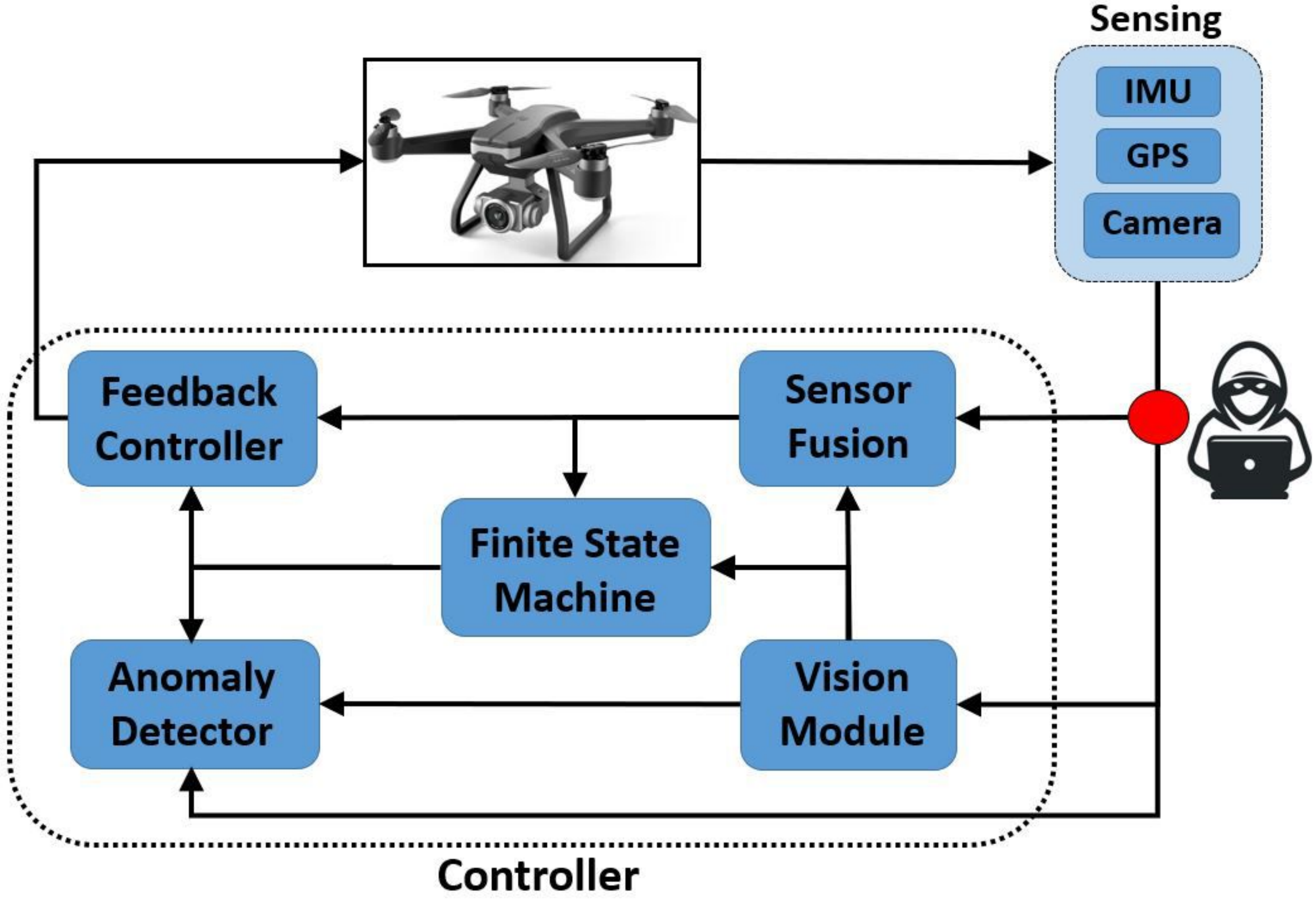}}
\vspace{-4pt}
\caption{The architecture of the perception-based UAV control system under attack on system sensing, including perception. Independently of the way attacks are actually implemented (e.g., directly compromising a sensor or modifying the measurements delivered over the network to the controller), the same impact on the control performance is obtained.} 
\label{fig:architecture}
\vspace{-6pt}
\end{figure}

We consider the common system architecture illustrated in Fig.~\ref{fig:architecture}
where raw sensor measurements from the IMU and GPS are used by a sensor fusion module 
to estimate the systems state,
whereas the relative position of the ground vehicle with respect to the camera is obtained by a vision module.  Depending on current state of the system, the Finite State Machine (FSM) 
switches between the drone control tasks, 
for VTOL and GVT as in~e.g.,~\cite{paris2020dynamic,wenzel2011automatic}.

We also assume that the system is equipped with an anomaly detector that utilizes the sensing information, as well as the controller and FSM outputs, to detect presence of abnormal behaviours. 
Further, we assume that an attacker can exploit the vulnerabilities of the employed communication network to compromise the sensing information delivered to the controller. 
In this work, we show that under such assumption, the attacker can design a sequence of false image and sensor values that can degrade the control performance 
while remaining stealthy from the employed anomaly detector, independent of the type of the utilized detector.

In what follows, we first describe the UAV's dynamical model that the attacker uses to generate the false images and sensor values at runtime. Then, we present the deployed vision-based controller and anomaly detector. Finally, the notion of stealthy and effective attacks is introduced as well as a methodology to design such attacks.

\subsection{Vehicle Dynamical Model}

The quadcopter's dynamical model can
be described using the standard Newton-Euler equations as
\begin{equation}\label{eq:plant}
\begin{split}
\dot{\mathbf{p}} &= \mathbf{v},\qquad\,\,
m\ddot{\mathbf{p}} = \mathbf{R}\mathbf{f}^{\mathcal{B}}-mg{{e}}_3,\\
\dot{\mathbf{R}}&=\mathbf{R}\hat{\bm{\omega}}^{\mathcal{B}},\quad
\bm{\tau}^{\mathcal{B}}=\bm{\mathcal{I}}\dot{\bm{\omega}}^{\mathcal{B}}+\bm{\omega}^{\mathcal{B}}\times \bm{\mathcal{I}}{\bm{\omega}}^{\mathcal{B}};
\end{split}
\end{equation}
here, $\mathbf{p}=\begin{bmatrix}x,y,z\end{bmatrix}^T$ and $\mathbf{v}$ denote the position and velocity of the vehicle in the earth frame $\mathcal{E}$, $m$ is the vehicle's mass, $\mathbf{R}\in\mathbb{R}^{ 3\times 3}$ is the rotation matrix from the body frame to the earth frame, $g$ is the gravity acceleration, $\mathbf{f}^{\mathcal{B}}=\begin{bmatrix}0,0,F_z\end{bmatrix}^T$ is the force vector in the body frame, ${e}_3=\begin{bmatrix}0,0,1\end{bmatrix}^T$ is the unit vector, $\bm{\omega}^{\mathcal{B}}=\begin{bmatrix}p,q,r\end{bmatrix}^T$ is the angular velocity in body frame, $\hat{.}$ denotes the operator that maps a vector in $\mathbb{R}^3$ to a skew-symmetric matrix, $\bm{\mathcal{I}}\in \mathbb{R}^{3\times 3}$ is the inertia matrix, and
$\bm{\tau}^{\mathcal{B}}=\begin{bmatrix}\tau_x,\tau_y,\tau_z\end{bmatrix}^T$ 
is the total torque vector in the body frame. 
The inputs to the 
system are the squared angular velocity of the four rotors $\mathbf{u}=\begin{bmatrix}w_1^2,w_2^2,w_3^2,w_4^2\end{bmatrix}^T$. The torques and thrust are obtained using the angular velocity of the four rotors as
\begin{equation}
\begin{bmatrix}F_z\\\tau_x\\\tau_y\\\tau_z\end{bmatrix}=\begin{bmatrix}b & b & b & b\\0 & -bl & 0 &bl\\-bl & 0 & bl &0\\d & -d & d & -d \end{bmatrix}\begin{bmatrix}w_1^2\\w_2^2\\w_3^2\\w_4^2\end{bmatrix},
\end{equation}
where $l$ is the distance from the motor to the center of gravity, and $b$ and $d$ are the thrust and drag coefficients,~respectively. 

\begin{figure}[!t]
\centerline{\includegraphics[width=\columnwidth]{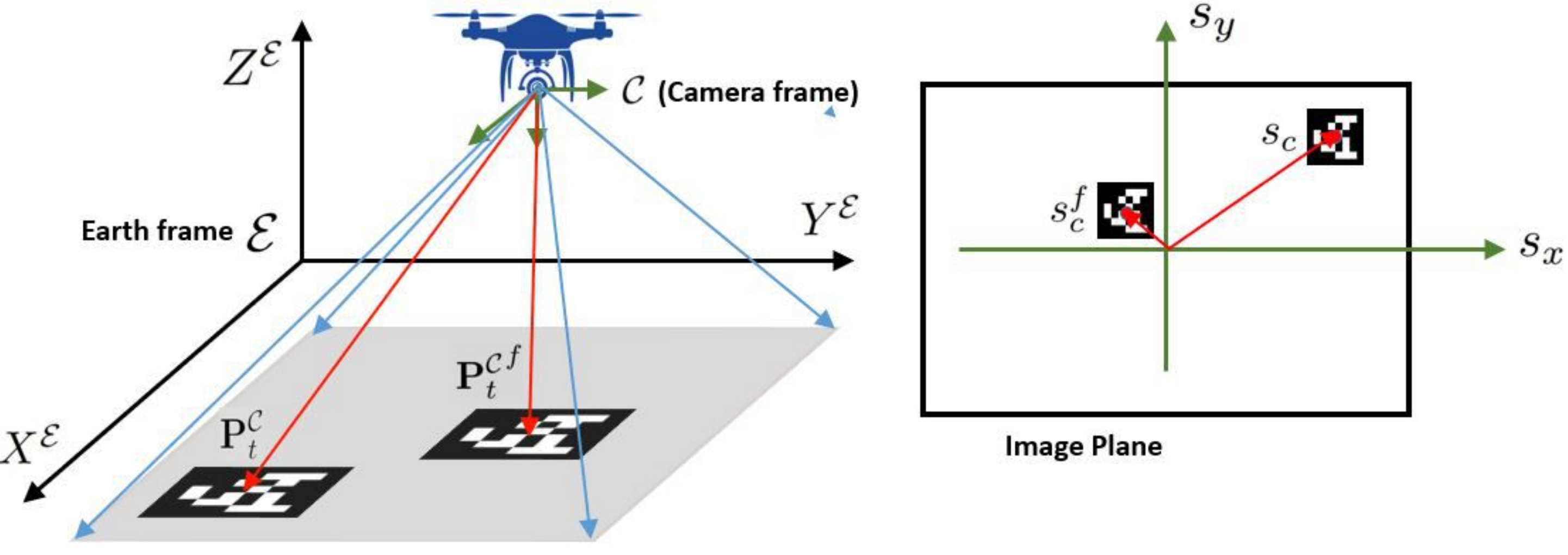}}
\vspace{-4pt}
\caption{Illustration of the drone with the landing mark, camera frame $\mathcal{C}$, earth frame $\mathcal{E}$, and image plane.} 
\label{fig:attack_image}
\vspace{-6pt}
\end{figure}

By defining $\mathbf{x}=\begin{bmatrix}x,y,z,\dot{x},\dot{y},\dot{z},\phi,\theta,\psi,p,q,r\end{bmatrix}^T$, after discretization the system~\eqref{eq:plant} can be captured in the state-space form
\vspace{-6pt}
\begin{equation}\label{eq:sys_cont}
\begin{split}
\mathbf{x}_{t+1} &= f(\mathbf{x}_{t},\mathbf{u}_{t})+\mathbf{w}_{t},\\
\mathbf{y}_t&= h(\mathbf{x}_t)+\mathbf{v}_t, \quad \mathbf{z}_{t} = G(\mathbb{X}_{t}^{\mathcal{C}}),
\end{split}
\end{equation}
where $\mathbf{w}_{t}$ is the system disturbance, $\mathbf{y}_{t}$ is the vector of the raw sensor measurements with Gaussian noise $\mathbf{v}_{t}$, $G$ is the  perspective camera projection model that maps the points represented in the camera frame $\mathcal{C}$ at time $t$, denoted by $\mathbb{X}_{t}^{\mathcal{C}}$, to the pixel images $\mathbf{z}_{t}$. Since the 3D points captured by camera $\mathbb{X}_{t}^{\mathcal{C}}$ are a function of the position of the camera and the UAV in the earth frame $\mathcal{E}$, $\mathbf{z}_t$ is explicitly a function of state $\mathbf{x}_{t}$; the corresponding frames are illustrated in Fig.~\ref{fig:attack_image}.

\subsection{Vision-Based Controller}
The objective of the vision-based controller is to use the camera-feed (i.e., images) and sensor measurements to compute the desired control inputs. Here, we utilize the widely-employed vision-based control framework as in e.g.,~\cite{araar2017vision,paris2020dynamic,lin2021low}. We consider a
squared fiducial marker ArUco~\cite{garrido2014automatic} as a landing mark. Specifically, a pin-hole model is used to find the relative position of the landmark center with respect to the camera frame (Fig.~\ref{fig:attack_image}, on the right) 
-- i.e., 
\begin{equation}\label{eq:pinhole}
s_c=\begin{bmatrix}s_{c_x}\\s_{c_y}\end{bmatrix}=\frac{f_c}{Z^{\mathcal{C}}}\begin{bmatrix}X^{\mathcal{C}}\\Y^{\mathcal{C}}\end{bmatrix},
\end{equation}
where $\begin{bmatrix}X^{\mathcal{C}}, Y^{\mathcal{C}}, Z^{\mathcal{C}}\end{bmatrix}^T \in\mathbb{X}_{t}^{\mathcal{C}}$ is a point on a 3D object visible to the pinhole camera in the camera frame $\mathcal{C}$ and $s_c=\begin{bmatrix}s_{c_x}&s_{c_y}\end{bmatrix}^T$ is the projected point on the image plane  with the focal distance~$f_c$ (see Fig.~\ref{fig:attack_image}). 

After detecting the landing mark position $s_c$ in the current image (Fig.~\ref{fig:attack_image}), using~\eqref{eq:pinhole} and the fact that the target (i.e., marker) size is known (which enables computing the $Z^\mathcal{C}$ in~\eqref{eq:pinhole}), 
the relative position of the landing mark \emph{with respect to the camera reference frame}, denoted by ${\mathbf{P}}_{t}^{\mathcal{C}}$, can be estimated from the image $\mathbf{z}_{t}$; i.e., we can define
\begin{equation}\label{eq:vision_m}
V(\mathbf{z}_{t})=\hat{\mathbf{P}}_{t}^{\mathcal{C}}={\mathbf{P}}_{t}^{\mathcal{C}}+\mathbf{e}({\mathbf{P}}_{t}^{\mathcal{C}}),
\end{equation}
where $V$ is the described mapping from an image $\mathbf{z}_t$ into $\hat{\mathbf{P}}_{t}^{\mathcal{C}}$, the estimated value of the  relative position of the landing mark in the camera frame (i.e.,  ${\mathbf{P}}_{t}^{\mathcal{C}}$).  
In addition, $\mathbf{e}({\mathbf{P}}_{t}^{\mathcal{C}})$ is the error of relative position estimation caused by the vision module and it is a function of the actual value of relative position ${\mathbf{P}}_{t}^{\mathcal{C}}$. 
Here, we use a common assumption for perception-based control (e.g.,~\cite{dean2021certainty}) that the norm of the camera-based localization error is bounded by some $\gamma >0$; i.e., $\Vert \mathbf{e}({\mathbf{P}}_{t}^{\mathcal{C}})\Vert \leq \gamma$. 

To connect the target position ${\mathbf{P}}_{t}^{\mathcal{C}}$ in the camera frame to the UAV state, in particular the drone position $\mathbf{p}_t$ 
captured in the earth (i.e., global) frame (see~\eqref{eq:plant}),  ${\mathbf{P}}_{t}^{\mathcal{C}}$ 
can be represented using the coordinates in the global frame as
\begin{equation}\label{eq:lm_incam}
{\mathbf{P}}_{t}^{\mathcal{C}}=\mathbf{R}_{\mathcal{E}}^{\mathcal{C}}(\mathbf{x}_{t})\Big({\mathbf{P}}_{t}^{\mathcal{E}}-\mathbf{p}_t\Big);
\end{equation}
here, $\mathbf{R}_{\mathcal{E}}^{\mathcal{C}}(\mathbf{x}_{t})$ is the rotation matrix from the earth frame $\mathcal{E}$ to the camera frame $\mathcal{C}$ and ${\mathbf{P}}_{t}^{\mathcal{E}}$ is the landing mark position in the earth frame $\mathcal{E}$. 

The UAVs 
control goals are determined by the 
FSM unit. 
Specifically, 
there are two states for 
the GVT; the first 
is to take off to the highest altitude such that the target vehicle is visible and can be detected. Then, the second 
is to follow the ground target vehicle at a certain altitude. For the VTOL task, there are three states. The first 
is similar to the first state of GVT. However, in the second state, the goal is to minimize the relative 
distance 
to the landing mark. Finally, if the relative 
distance from the target is less than a predefined threshold, the drone switches to the \emph{landing-mode}, which is the final state for VTOL.
%
For each VTOL and GVT~task, a cascade PID controller uses $\hat{\mathbf{P}}_{t}^{\mathcal{C}}$ and the sensor fusion output to control the position and the attitude of the drone. Thus, we assume that the closed-loop control system~is~stable.

%
%

\subsection{Anomaly Detector}
Physics-based anomaly detectors including (but not limited to) $\chi^2$, cumulative sum (CUSUM), sequential probability ratio test (SPRT) have been widely deployed in robotics systems like UAVs (e.g.,~\cite{jovanov_tac19,kwon2014analysis}).
Recently, 
learning-based anomaly detectors 
have been employed for detecting the presence of attacks or any system anomalies (e.g.,~\cite{kravchik2018detecting}).
Without loss of generality, assume that the attack starts at time $t=0$, and let us use  $\mathbf{O}_t=\{\mathbf{z}_{t},\mathbf{y}_{t}\}$ 
and 
$\mathbf{O}^a_t=\{\mathbf{z}^a_{t},\mathbf{y}^a_{t}\}$
to respectively denote the real and compromised (i.e., false) sensing measurements at time~$t$, which includes the sensor values and the camera image. 
Now, given a random sequence $\bar{\mathbf{O}}^t = \{\bar{\mathbf{O}}_{-T_0}:\bar{\mathbf{O}}_t\}$ of the sensing data received by the controller,  any employed detector 
(no matter~the~specific~detector design)
is effectively trying to solve the hypothesis problem: 

\vspace{4pt}$H_0$:  normal behavior---i.e., $\mathbf{O}_{-T_0}:\mathbf{O}_t$ was  received; 

$H_1$: abnormal behavior---i.e., $\mathbf{O}_{-T_0}:\mathbf{O}_{-1},\mathbf{O}_{0}^a:\mathbf{O}_t^a$ was received.\vspace{4pt}
\\
Thus, the sequence $\bar{\mathbf{O}}^t$ 
either comes from the distribution of the null hypothesis $H_0$, which is determined by the system uncertainties, or from a distribution of the alternative hypothesis $H_1$, \emph{which is unknown but controlled by the attack signals}. For a given anomaly detector specified by a function $D: \bar{\mathbf{O}}^t\to \{0,1\}$, false alarm occurs if $D(\bar{\mathbf{O}}^t)=1$ when $\bar{\mathbf{O}}^t$ comes from $H_0$, and we denote the probability of false alarm as $p^{FA}(D)$;  whereas true detection occurs if $D(\bar{\mathbf{O}}^t)=1$ when $\bar{\mathbf{O}}^t$ comes from $H_1$, and we denote the probability of true detection as $p^{TD}(D)$.


\section{Design of Stealthy Effective Attacks}
\label{sec:attacks}

In this section, we introduce an algorithm to  attack consistently both the images and the sensors, ensuring both attack effectiveness and stealthiness. 
We assume that the attacker has access to the current sensor measurements and camera images (i.e., $\mathbf{z}_{t},\mathbf{y}_{t}$).
Now, to formally capture the attacker's goal, 
we start by formalizing the stealthiness notion and the attack impact on the control performance. 


\begin{definition}\label{def:stealthy}
An attack is \emph{{strictly stealthy}} if 
for a resulting observation sequence $\bar{\mathbf{O}}^t$,
there exists no detector for which $p_t^{FA}<p_t^{TD}$ holds, for any $t\geq 0$. An attack is
\emph{$\epsilon$-{stealthy}} if for 
a resulting observation sequence $\bar{\mathbf{O}}^t$
and 
a given $\epsilon >0$, there exists no detector such that $p_t^{FA}<p_t^{TD}-\epsilon$ holds, for any $t\geq 0$. 
\end{definition}

From the definition, the attack sequence is strictly stealthy if there is no detector for which the probability of true detection is greater than the probability of false alarm. However, reaching that level of stealthiness may not be possible 
and therefore, we also define $\epsilon$-{stealthiness} for which the difference between the probabilities of true detection and false alarm is less than $\epsilon$ for any employed detector. 

Since the goal of the vision-based control is to follow the target vehicle, the attack degrades control performance by causing a deviation from the desired trajectory of the drone in the 3-D space.  
In particular, the attack goal is to force the drone away from the target vehicle by some $\alpha$; this is equivalent to moving the target landing marker in \emph{the camera frame} (i.e., $\mathbf{P}_t^{\mathcal{C}}$) $\alpha$-meters away (in 3-D space). 
Thus, we introduce the following definition.
\begin{definition}
An attack is \emph{$\alpha$-effective} if $\Vert \mathbf{P}_t^{\mathcal{C}}\Vert_2 \geq \alpha$ for some $t\geq 0$. 
\end{definition}
%

Now, 
%
the attack goal can be formalized as designing a sequence of false sensing information from time $t=0$, i.e., $\{\mathbf{y}_{0}^f,\mathbf{z}_{0}^f\},...,\{\mathbf{y}_{T}^f,\mathbf{z}_{T}^f\}$, that is 
$\alpha$-effective and $\epsilon$-stealthy 
for all $t\in [0,T]$. A sequence of false sensing values that satisfy the attack goal is referred to as an \emph{$(\epsilon,\alpha)$-attack sequence}.

\subsection{Attack Design}
The previous works~\cite{khazraei_l4dc22,khazraei2022attack,mo2010false,khazraei2022resiliency,khazraei2022attacks,hallyburton2022optimal} have shown that stealthiness conditions require that the attack impact (i.e., control degradation) is compatible with the system dynamics. For example, if the attacker wants to cause deviation in the drone trajectory 
(compared to the unattacked system's trajectory), the deviation needs to increase smoothly and in accordance with the dynamics of the system. 

Specifically, if $\mathbf{s}_{t}$ 
is used to denote the state deviation 
due to the attack, 
to satisfy the stealthiness  condition,~\cite{khazraei_l4dc22,khazraei2022resiliency} show that $\mathbf{s}_t$ needs to dynamically evolve over time as follows 
\begin{equation}
\label{eq:temp1}
\mathbf{s}_{t+1} = f(\mathbf{\hat{x}}_{t}^a,\mathbf{u}_{t})-f(\mathbf{\hat{x}}_{t}^a-\mathbf{{s}}_{t},\mathbf{u}_{t}),
\end{equation}
where $\mathbf{s}\in \mathbb{R}^{12}$ and $f$ is defined in~\eqref{eq:sys_cont}, with some small nonzero initial condition $\mathbf{s}_{0}$; also, $\mathbf{\hat{x}}_{t}^a$ denotes the output of a sensor fusion that the attacker runs online using the sensing information at each time step 
(e.g., using $\mathbf{y}_{t}$, and potentially $\mathbf{z}_t$). 
Note that $\mathbf{\hat{x}}_{t}^a$ differs from the system's sensor fusion output $\mathbf{\hat{x}}_{t}$, as the former is obtained from the actual sensing measurements  whereas the latter is computed 
using the compromised sensing information (i.e., 
$\mathbf{y}_{t}^f)$. 

We next show how the attacker 
can exploit the sequence $\mathbf{s}_t$, $t\geq 0$, to design a stealthy and effective sequence of false physical sensor and image data.

\subsubsection*{Attacks on Sensor Measurements}
From~\eqref{eq:sys_cont}, 
the physical sensors map the states $\mathbf{x}_t$ to the observation values with function $h$. Thus, to remain stealthy, the false sensor values should be obtained by mapping $\mathbf{x}_t-\mathbf{s}_t$ using function $h$ -- i.e., 
ideally, the falsified sensor values should be $\mathbf{y}_{t}^f=h(\mathbf{x}_t-\mathbf{s}_t)+\mathbf{v}_t$.
However, 
there are two challenges to construct such falsified sensor data: 
no access to the actual state of the system $\mathbf{x}_t$ and measurement noise $\mathbf{v}_t$. 

For the former, the estimated state value $\hat{\mathbf{x}}_{t}^a$ can be used instead of $\mathbf{x}_t$. 
For the latter, the attacker can subtract $h(\mathbf{\hat{x}}_{t}^a)$ from the current sensor measurements $\mathbf{y}_{t}$ to estimate the noise level. Combining these, 
the sequence of false physical sensor values for $t\geq 0$ 
can be computed as
\begin{equation}
\label{eq:sense_attack}
\mathbf{y}_{t}^f=\mathbf{y}_{t}+\mathbf{a}_{t}= 
\mathbf{y}_{t}+h(\mathbf{\hat{x}}_{t}^a-\mathbf{{s}}_{t})-h(\mathbf{\hat{x}}_{t}^a);
\end{equation}
i.e., the attacker just needs to add $\mathbf{a}_{t}=h(\mathbf{\hat{x}}_{t}^a-\mathbf{{s}}_{t})-h(\mathbf{\hat{x}}_{t}^a)$ to the current physical sensor measurement $\mathbf{y}_{t}=h(\mathbf{x}_t)+\mathbf{v}_t$. 
As the state estimation error $\mathbf{{x}}_{t}-\mathbf{\hat{x}}_{t}^a$ 
decreases, $\mathbf{y}_{t}^f$ from~\eqref{eq:sense_attack} approaches the ideal (from the perspective of the attack effectiveness and stealthiness) false sensor value -- i.e., $\mathbf{y}_{t}^f=h(\mathbf{x}_t-\mathbf{s}_t)+\mathbf{v}_t$.

\subsubsection*{Attacks on Camera Images}
Similarly, to remain stealthy, the attacker also needs to remove from the actual image, the impact of the deviation caused by the attack. To achieve this, our approach is to first identify the desired false position and size of the landing mark in the image plane. 
This is done by finding the desired false position of the landing mark in the camera frame and then using~\eqref{eq:pinhole} to get the desired false position and size of the landing mark in the image plane (see for example Fig.~\ref{fig:attack_image}). 

When the drone is not under attack, the current position of the landing mark in the camera frame (i.e., ${\mathbf{P}}_{t}^{\mathcal{C}}$) satisfies~\eqref{eq:lm_incam}. 
To satisfy the stealthiness constraint by removing the deviation, the idea is to replace all the system states in~\eqref{eq:lm_incam} with $\hat{\mathbf{x}}_{t}^a-\mathbf{s}_t$; i.e., 
the false position of the landing mark in the camera frame, denoted by  ${\mathbf{P}}_{t}^{\mathcal{C}}$, should be computed as
\begin{equation}\label{eq:attack_image}
{{\mathbf{P}}_{t}^{\mathcal{C}}}^f=\mathbf{R}_{\mathcal{E}}^{\mathcal{C}}(\hat{\mathbf{x}}_{t}^a-\mathbf{s}_t)\Big(\hat{\mathbf{P}}_{t}^{\mathcal{E}}-(\hat{\mathbf{p}}_t^a-\mathbf{s}^{\mathbf{p}}_t)\Big);
\end{equation}
here, $\mathbf{R}_{\mathcal{E}}^{\mathcal{C}}(\hat{\mathbf{x}}_{t}^a-\mathbf{s}_t)$ is the rotation matrix from the earth frame to the camera frame evaluated by fake Euler angles associated with $\hat{\mathbf{x}}_{t}^a-\mathbf{s}_t$; $\hat{\mathbf{P}}_{t}^{\mathcal{E}}$ is the attacker's estimate of the position of the landing mark in the earth frame; 
$\mathbf{s}^{\mathbf{p}}_t$ captures the first three elements of the vector $\mathbf{s}_t$ (associated with the vector $\mathbf{p}=\begin{bmatrix}x,y,z\end{bmatrix}^T$). Note that to estimate $\hat{\mathbf{P}}_{t}^{\mathcal{E}}$, the attacker can either have its own resources (e.g., placing sensing such as GPS on the target and estimating the position using a Kalman filter) or can use the actual image $\mathbf{z}_t$ to find the relative position and then combine it with the drone position in the earth frame $\mathcal{E}$ (the first three elements of $\hat{\mathbf{x}}_{t}^a$).

\begin{algorithm}[!t]
  \caption{Design of stealthy and effective attacks by compromising both physical sensors and image starting at time 0 until $T$}
  \label{alg1}
  \begin{algorithmic}[1] 
  \State {Initialize $s_0$}
    \For{$t=0:T$}
        \State Find $\hat{\mathbf{x}}_{t}^a$ and $\hat{\mathbf{P}}_{t}^{\mathcal{E}}$
        \State $\mathbf{y}_{t}^f=\mathbf{y}_{t}+h(\mathbf{\hat{x}}_{t}^a-\mathbf{{s}}_{t})-h(\mathbf{\hat{x}}_{t}^a)$
        \State ${{\mathbf{P}}_{t}^{\mathcal{C}}}^f=\mathbf{R}_{\mathcal{E}}^{\mathcal{C}}(\hat{\mathbf{x}}_{t}^a-\mathbf{s}_t)\Big(\hat{\mathbf{P}}_{t}^{\mathcal{E}}-(\hat{\mathbf{p}}_t^a-\mathbf{s}^{\mathbf{p}}_t)\Big)$
        \State $s_{c}^f=\begin{bmatrix}s_{c_x}^f\\s_{c_y}^f\end{bmatrix}=\frac{f_c}{Z_{{{\mathbf{P}}_{t}^{\mathcal{C}}}^f}}\begin{bmatrix}X_{{{\mathbf{P}}_{t}^{\mathcal{C}}}^f}\\Y_{{{\mathbf{P}}_{t}^{\mathcal{C}}}^f}\end{bmatrix}$
        \State $l^f=\frac{f_c}{Z_{{{\mathbf{P}}_{t}^{\mathcal{C}}}^f}}l$
        \State 
        {Generate $\mathbf{z}_t^f$ with $s_{c}^f$ and $l^f$}
        \State
        $\mathbf{s}_{t+1} = f(\mathbf{\hat{x}}_{t}^a,\mathbf{u}_{t})-f(\mathbf{\hat{x}}_{t}^a-\mathbf{{s}}_{t},\mathbf{u}_{t})$
        
    \EndFor
  \end{algorithmic}
\end{algorithm}

Now, 
the attacker should design an image at each time $t\geq 0$ for which the vision module output~\eqref{eq:vision_m} results in ${{\mathbf{P}}_{t}^{\mathcal{C}}}^f$ from~\eqref{eq:attack_image}. Once ${{\mathbf{P}}_{t}^{\mathcal{C}}}^f$ is obtained, the desired false location for the landing mark center in the image plane can be found using~\eqref{eq:pinhole} 
-- i.e.,
\vspace{-8pt}
\begin{equation}\label{eq:center}
s_{c}^f=\begin{bmatrix}s_{c_x}^f\\s_{c_y}^f\end{bmatrix}=\frac{f_c}{Z_{{{\mathbf{P}}_{t}^{\mathcal{C}}}^f}}\begin{bmatrix}X_{{{\mathbf{P}}_{t}^{\mathcal{C}}}^f}\\Y_{{{\mathbf{P}}_{t}^{\mathcal{C}}}^f}\end{bmatrix}.
\end{equation}
On the other hand, the scale of the landing mark in the image plane needs to be compatible with the desired output of the vision module ${{\mathbf{P}}_{t}^{\mathcal{C}}}^f$. If the actual physical length of the squared ArUco marker is $l$, the length of the landing mark in the fake image plane $l^f$ can be obtained as 
\begin{equation}\label{eq:scale}
l^f=\frac{f_c}{Z_{{{\mathbf{P}}_{t}^{\mathcal{C}}}^f}}l.
\end{equation}
The resulting procedure is summarized in Algorithm~\ref{alg1}.

Finally, to find a falsified image that satisfies~\eqref{eq:center} and~\eqref{eq:scale}, 
different methods can be employed. 
A promising approach is to use deep learning based generative models such as conditional generative adversarial network (cGAN) or similar generative approaches~\cite{lambert2018deep}. In this work, we use common computer vision methods, as  described in the next~section.


%% file: Evaluation.tex
\section{Results and Evaluation}\label{sec:result}

We now describe the performed evaluation of the effectiveness and stealthiness of the introduced
attack methodology on the GVT and VTOL control tasks. 
Representative experimental videos are provided at~\cite{video}.

\subsection{Implementation Details}
\begin{figure}
  \begin{subfigure}[b]{0.23\textwidth}
    \includegraphics[width=\textwidth,height=3cm]{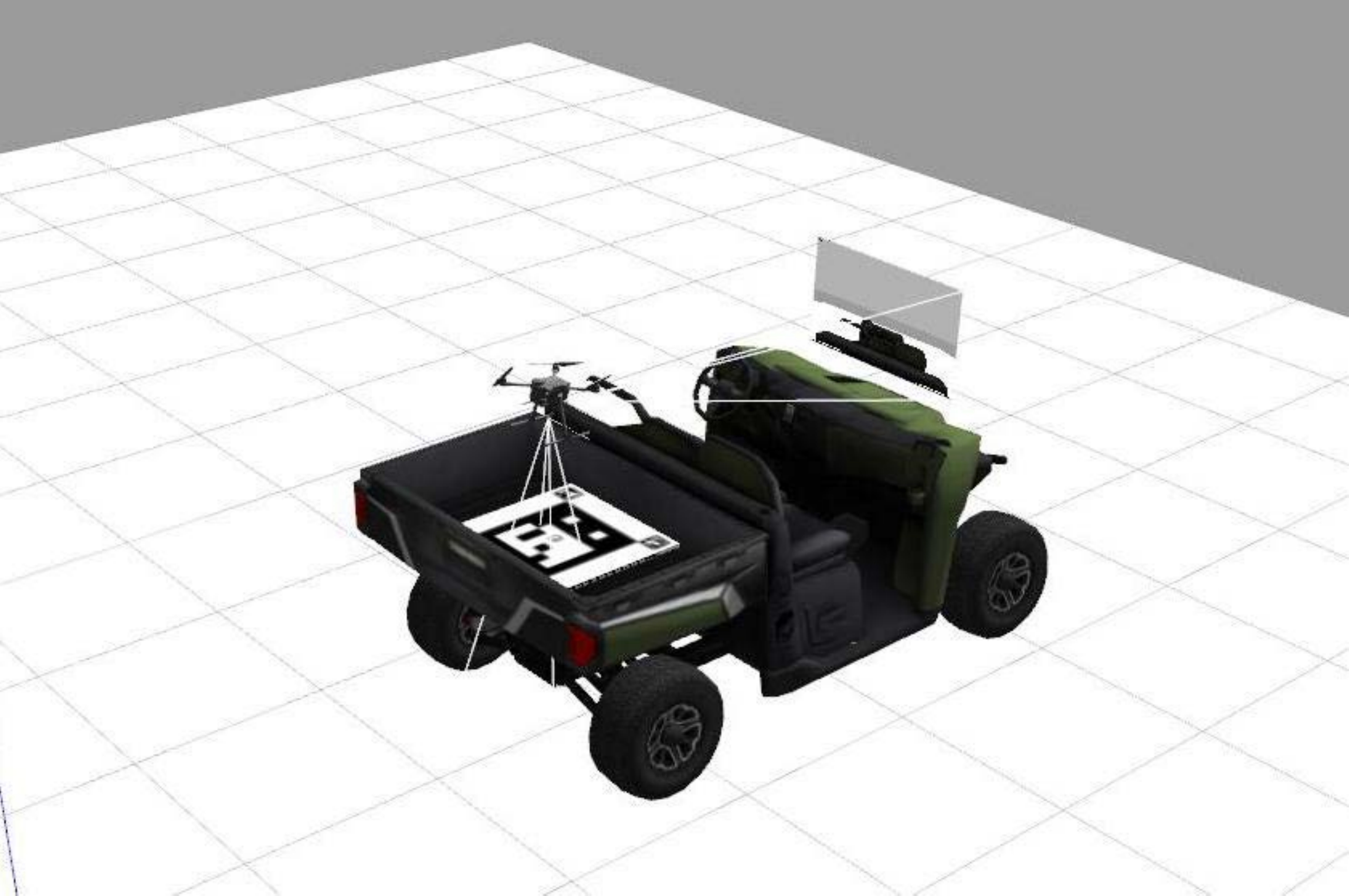}
  \end{subfigure}
  \begin{subfigure}[b]{0.23\textwidth}
    \includegraphics[width=\textwidth,height=3cm]{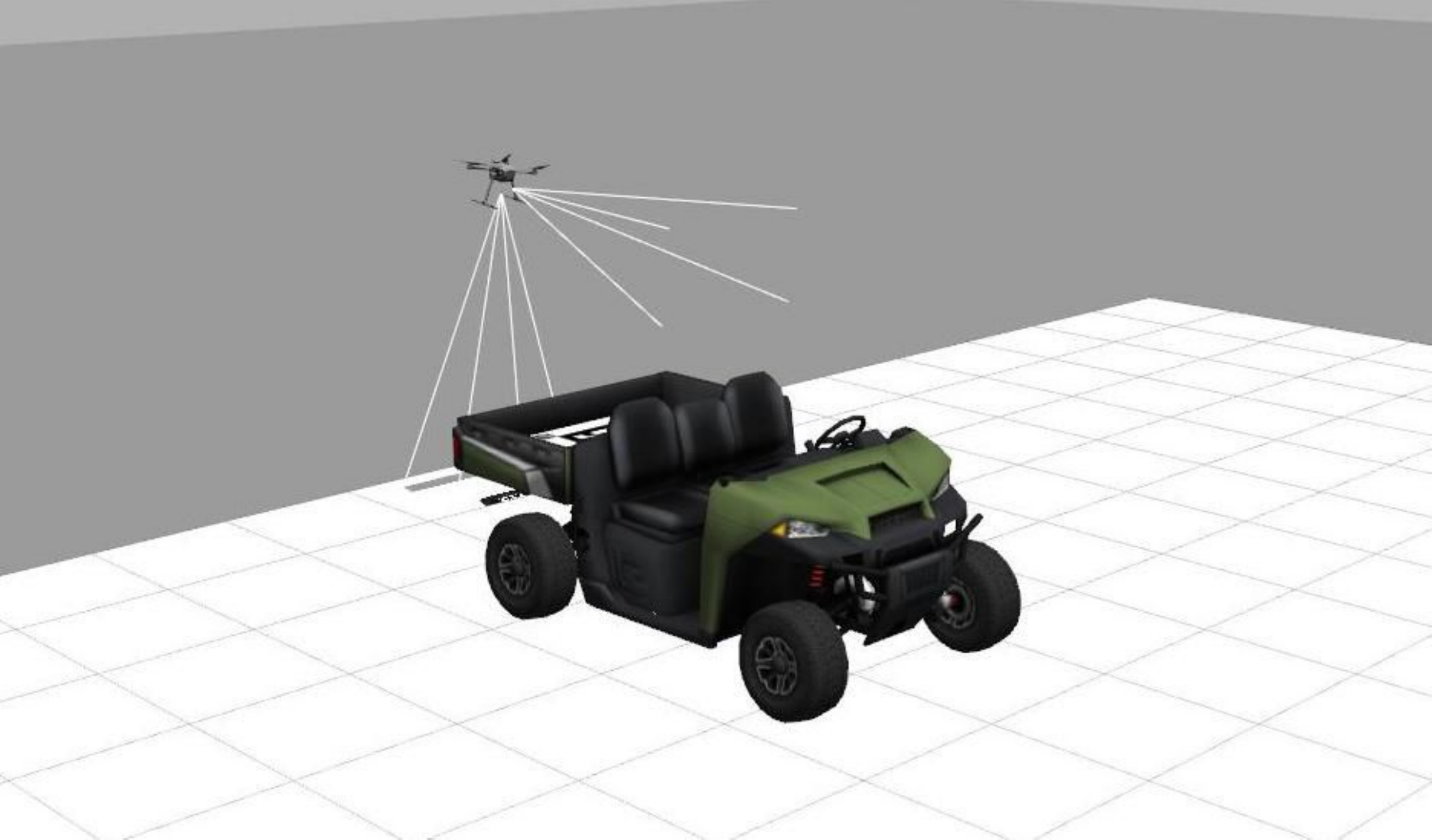}
  \end{subfigure}
  \vspace{-4pt}
  \caption{The simulation environment.} 
  \label{fig:snapshot}
  \vspace{-6pt}
\end{figure}

We 
evaluated the presented attacks using 
an open-source autonomous vehicle simulation platform Prometheus~\cite{Prometheus}
powered by the PX4 flight controller 
and Gazebo 3D robotics simulator. The built-in quadcopter uses a sensor fusion model of IMU and GPS and is equipped with a down-facing camera with a resolution of $1280\times720$, allowing us to implement the control architecture from Fig.~\ref{fig:architecture}.
%
In our simulations, the quadcopter identifies and tracks an ArUco marker placed on a ground vehicle to perform autonomous tracking and landing missions (as illustrated 
in Fig.~\ref{fig:snapshot}). 

Attacks on sensors were implemented using 
Algorithm~\ref{alg1}. To generate false images, we assumed that the attacker has a base image with similar background, moving 
the position of the ground vehicle 
with the landing marker and resizing them by up-sampling (or down-sampling) using the formulas in Algorithm~\ref{alg1}. Two different anomaly detectors were implemented to evaluate attack stealthiness: $\chi^2$ and learning-based detectors. $\chi^2$ detector uses a weighted norm of the residue generated by the Extended Kalman filter, raising 
alarm if the residue is larger than a predefined threshold. We also implemented a widely-used learning-based detector~\cite{kravchik2018detecting} that employs a long short-term memory (LSTM) architecture. 

\subsection{Attacks on Ground Vehicle Tracking}
\begin{figure*}[!t]
\begin{subfigure}{0.312\textwidth}
\includegraphics[width=.95\linewidth, height=4cm]{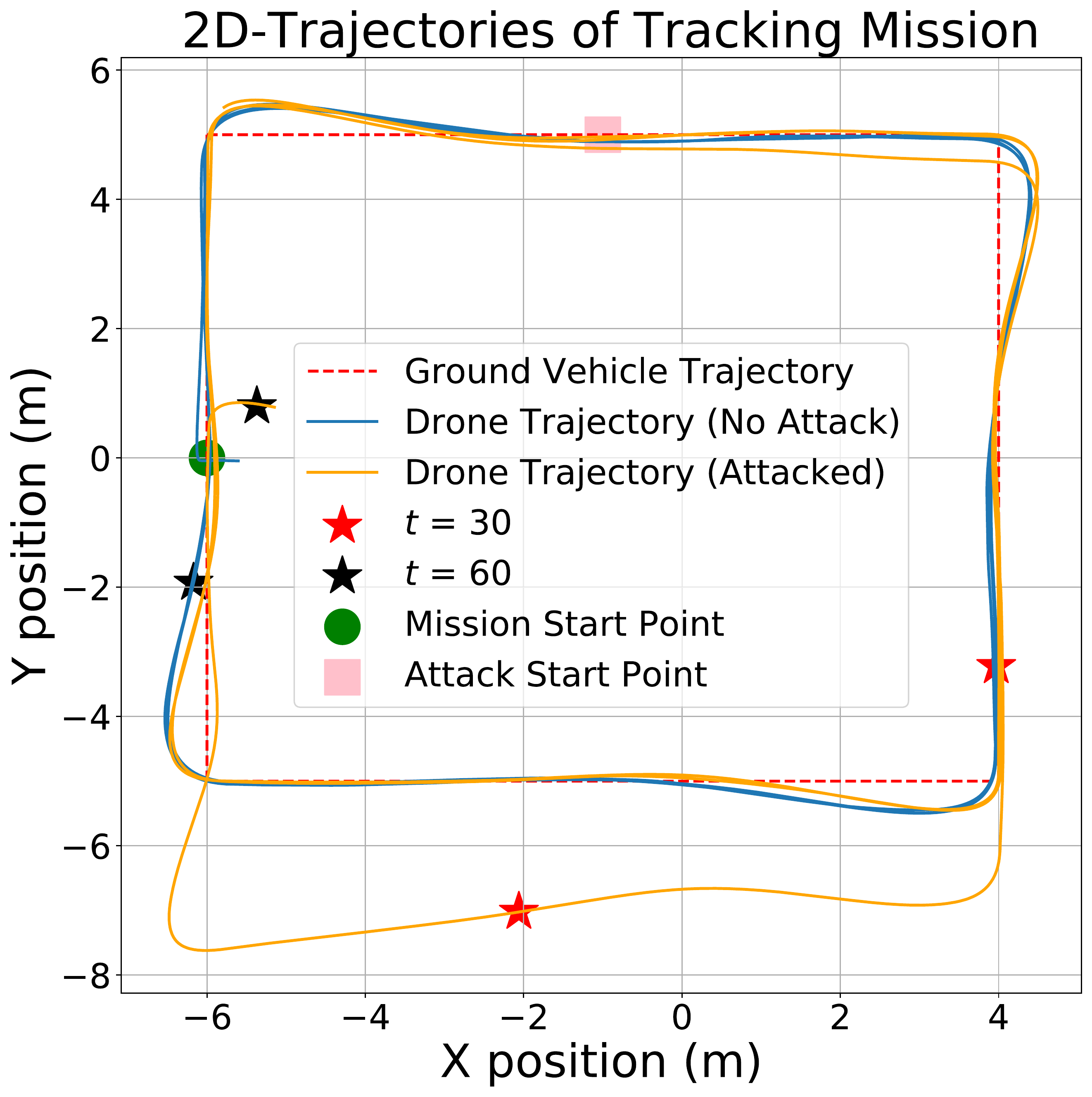}
\vspace{-6pt}
\caption{ }\label{fig:drone_fig1}
\end{subfigure}
\begin{subfigure}{0.312\textwidth}
\includegraphics[width=.95\linewidth, height=4cm]{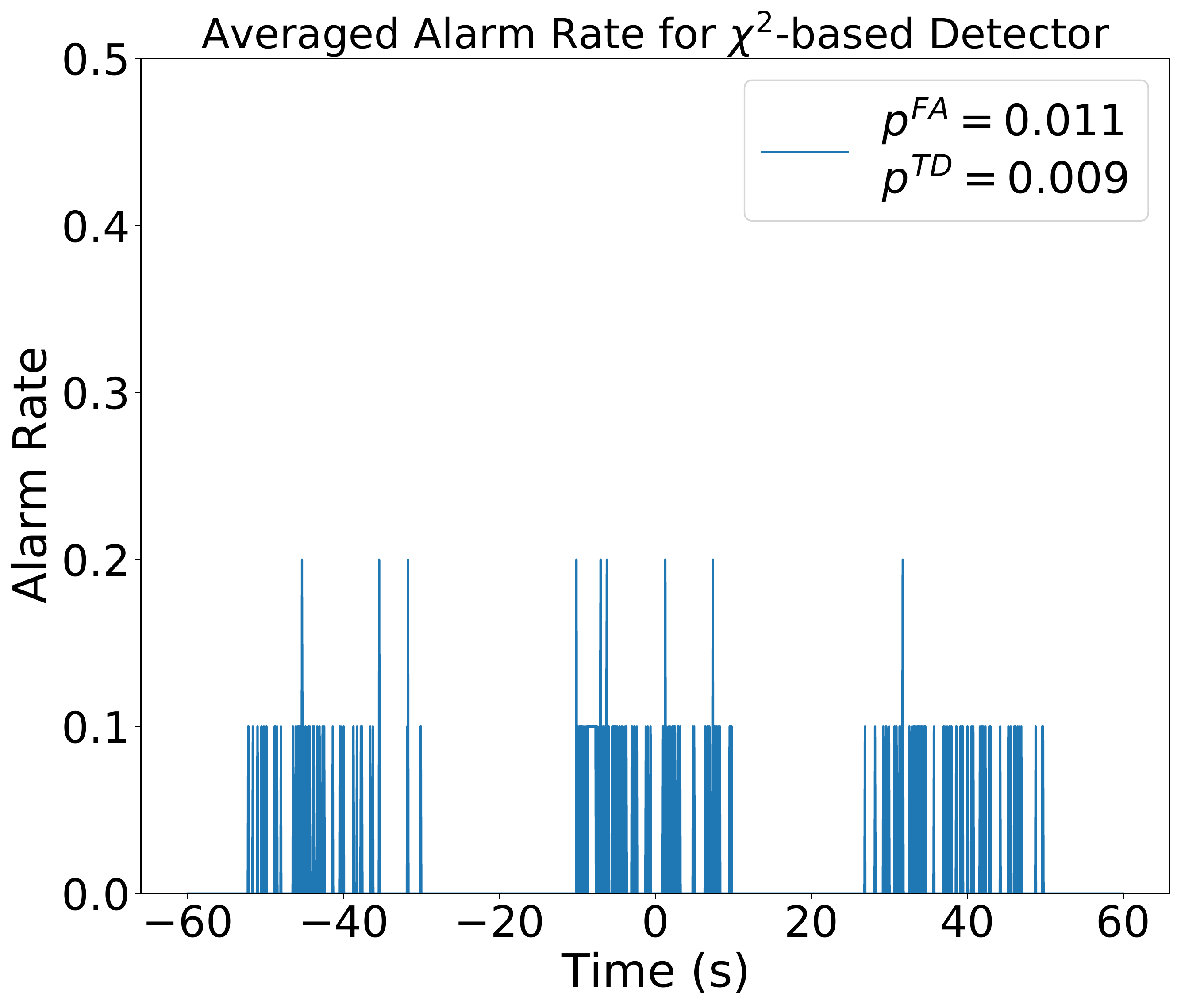}
\vspace{-6pt}
\caption{}\label{fig:drone_fig2}
\end{subfigure}
\begin{subfigure}{0.312\textwidth}
\includegraphics[width=.95\linewidth, height=4cm]{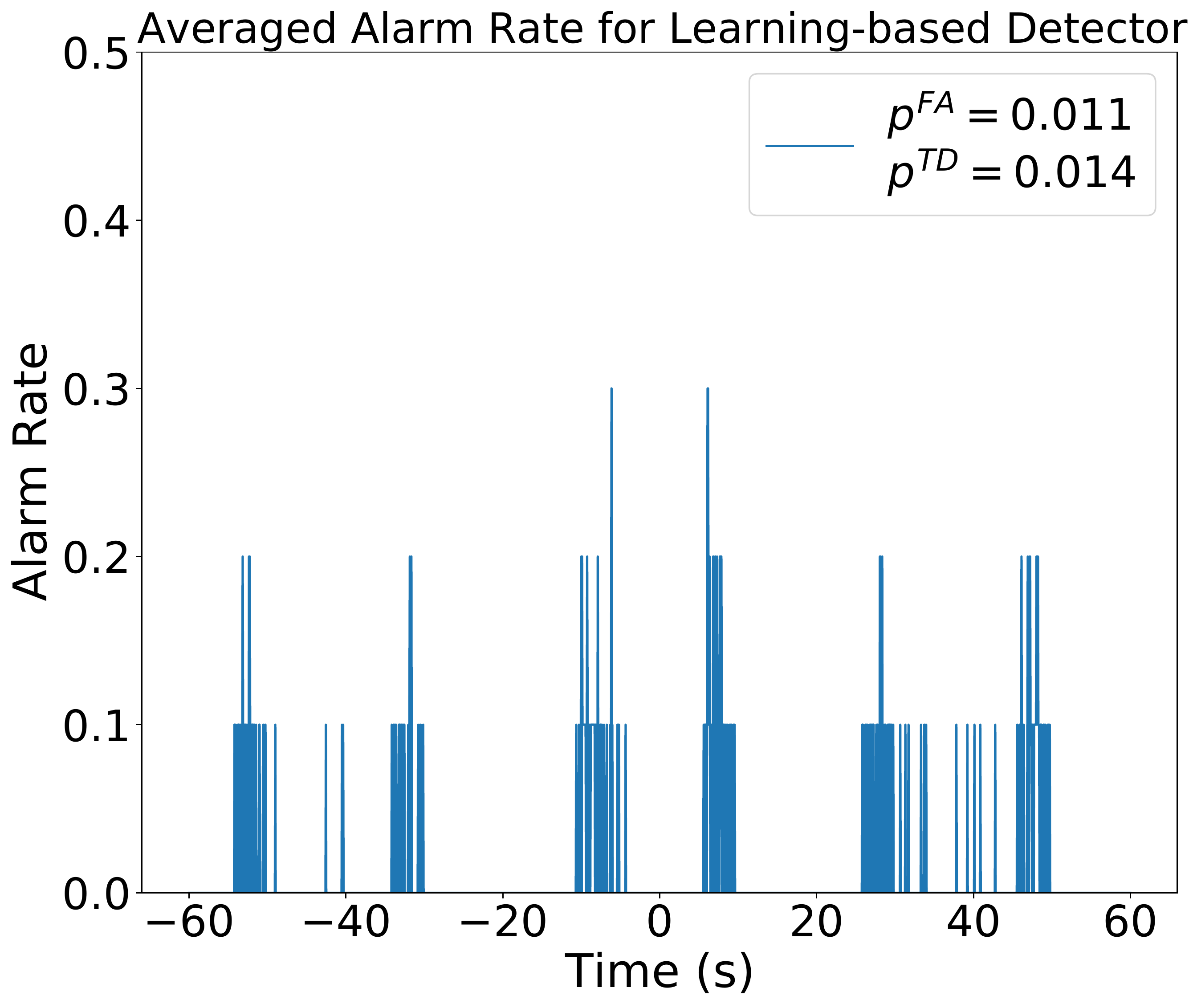}
\vspace{-6pt}
\caption{}\label{fig:drone_fig3}
\end{subfigure}
\begin{subfigure}{0.312\textwidth}
\includegraphics[width=.95\linewidth, height=4cm]{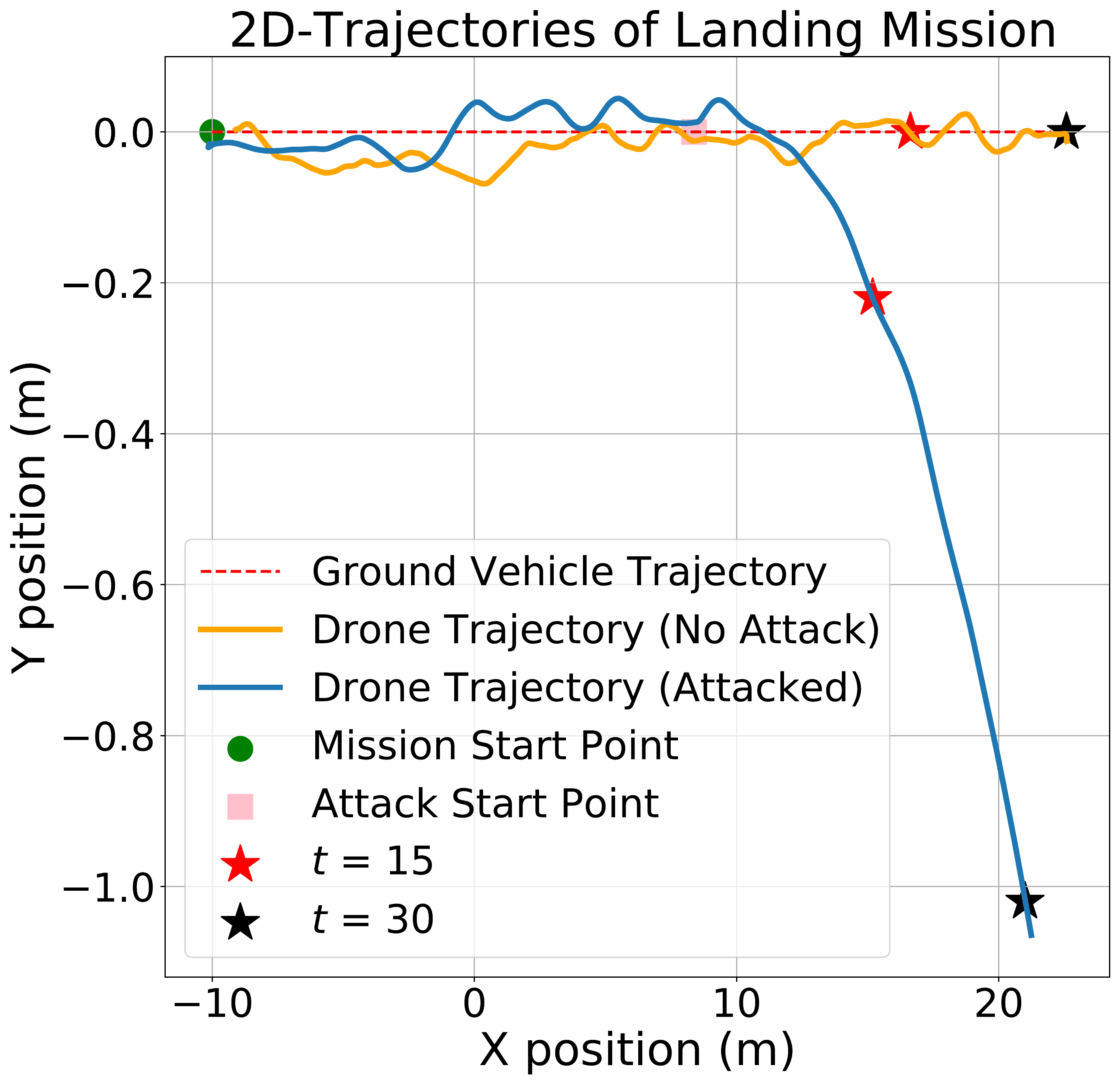}
\vspace{-6pt}
\caption{}\label{fig:drone_fig4}
\end{subfigure}
\begin{subfigure}{0.312\textwidth}
\includegraphics[width=.95\linewidth, height=4cm]{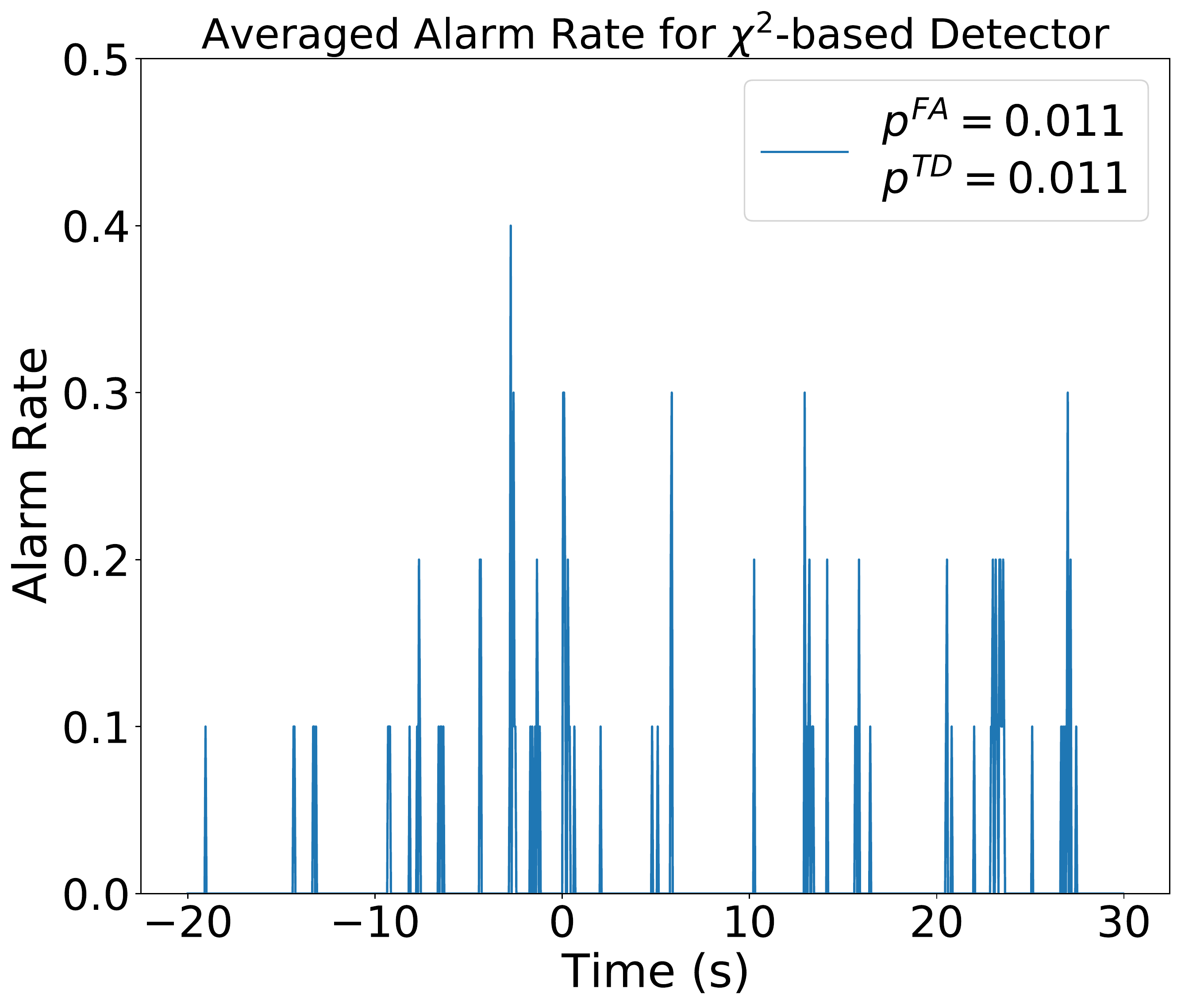}
\vspace{-6pt}
\caption{}\label{fig:drone_fig5}
\end{subfigure}
\begin{subfigure}{0.312\textwidth}
\includegraphics[width=.95\linewidth, height=4cm]{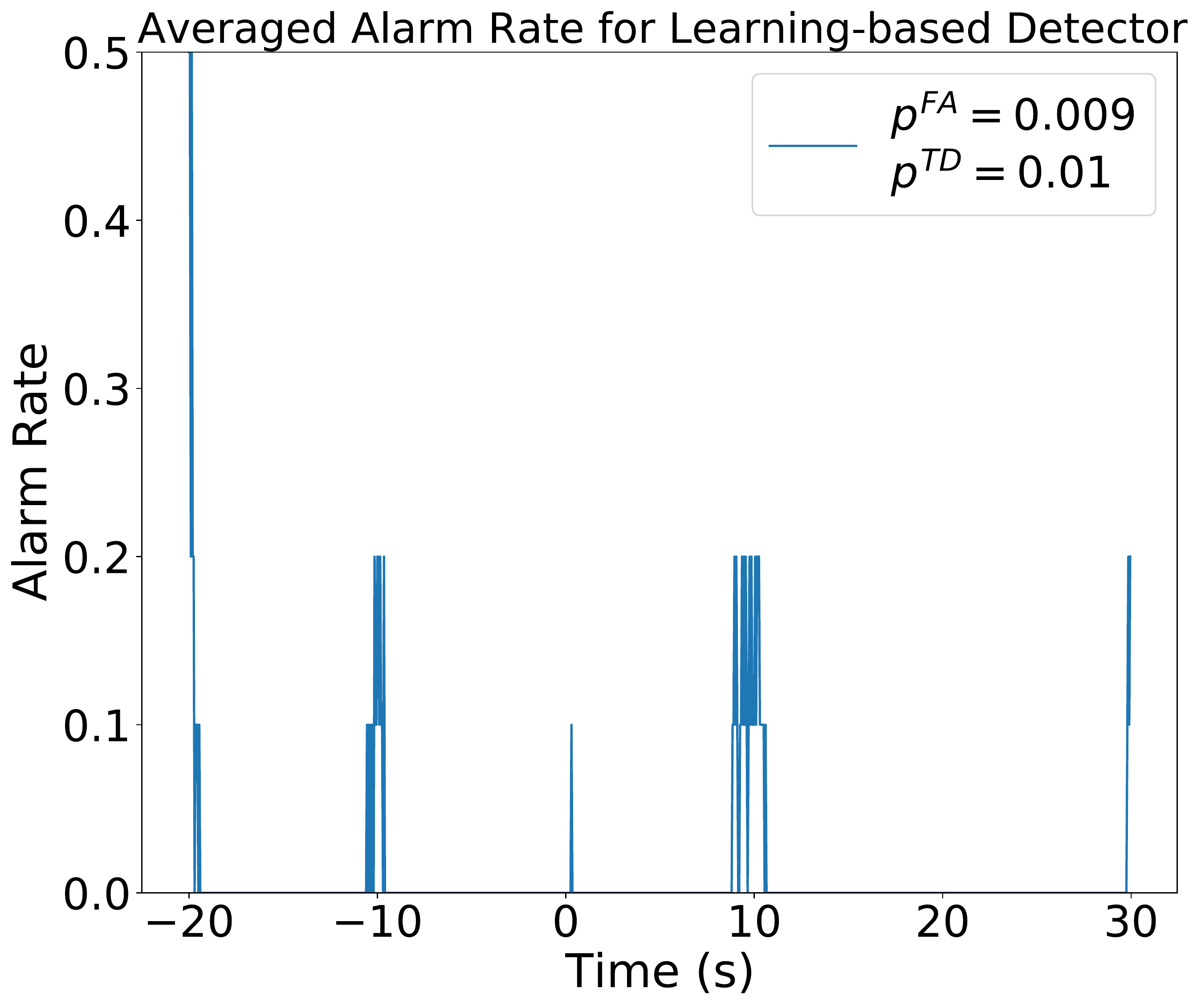}
\vspace{-6pt}
\caption{}\label{fig:drone_fig6}
\end{subfigure}
\vspace{-6pt}
\caption{(a)~The trajectories of the UAV and ground vehicle in $XY$ coordinates over time for the GVT. The blue and orange lines show the UAV trajectory without and during the attack, respectively; 
(b,c)~The average alarm rate for $\chi^2$ and learning-based anomaly detectors before and after the attack (attack starts at time $t=0$) for the GVT task over 10 experiments; (d)~The trajectories of the drone and the ground vehicle in $XY$ coordinate for VTOL; (e,f)~The alarm rate for $\chi^2$ and learning-based anomaly detectors before and after the attack (starting at $t=0$) for the VTOL task over~10~experiments.}
\label{fig:drone_fig}
\vspace{-8pt}
\end{figure*}

In this task, the drone's goal is to track the ground vehicle that follows a square trajectory clockwise 
illustrated in Fig.~\ref{fig:drone_fig1} with the red dashed line. The blue line shows the trajectory of the drone without attack, 
successfully tracking the vehicle. The orange trajectory shows the drone trajectory when the attack starts at the pink square. In this experiment, we assumed the  elements of the initial $\mathbf{s}_0$ were all zero except the one associated with the roll angle which was set to $0.01$. 
With such initial condition, the attacker was successful in deviating the drone trajectory along the $Y$ axis up to $3~m$ at which point the ground vehicle 
went out of the range of the camera's scope and we stopped the attack; note e.g., that red stars highlight the drone positions at the same times $t=30~s$ for trajectories under attack and without attack, illustrating significant deviations due to the attack.
Similar results were shown for other $\mathbf{s}_0$ initializations, where the error was along the $X$ axis, or the combination of both axes.

Fig.~\ref{fig:drone_fig2} and Fig.~\ref{fig:drone_fig3} show the average alarm rate at each time step for 10 experiments when both $\chi^2$ and learning-based anomaly detectors were used, respectively. As the attack started at time $t=0$, the value of true alarm averages (for $t > 0$) $p^{TD}=.01$  was almost the same as the false alarm averages (for $t\geq 0$) for both detector, showing the $\epsilon$-stealthiness of the attack according to Definition~\ref{def:stealthy}.

Table~\ref{tab:GVT_stealthy} illustrates the impact of the norm of the initial $\mathbf{s}_0$ 
on the stealthiness level. 
The true detection rate 
increases with the increase in the norm of the initial error state $\mathbf{s}_0$. Intuitively, this is 
caused by a huge initial error bias due to the attack at time $t=0$ (attack start), which can be easily detected by an anomaly detector. On the other hand, for $\mathbf{s}_0$ arbitrarily close to zero the attack will not be detected, but it will take longer time for the attack to be effective. Therefore, there is a trade-off between  the attack detectability and the time 
required to achieve significant performance degradation. 

So far, we assumed that the attack is implemented on both sensors and images consistently according to Algorithm~\ref{alg1}. However, Table~\ref{tab:Tracking_incon} shows the probability of attack detection for different values of deviation when the sensors were not under attack and only images were compromised -- 
even for a small deviation of $0.4~m$ the attack was detected by $\chi^2$ detector with true detection rate of $0.79$ which is 79 times higher than the false alarm rate ($p^{FA}=0.01$). Intuitively, this 
is caused by the inconsistency between the sensors and the images, with the sensor values being based on the true drone position whereas the attacked image were consistent with a different positioning value.


\begin{table}[!t]
    \centering
    \caption{Average true detection rate for $\chi^2$ and learning-based anomaly detectors for different values of the initial condition norm ($\Vert \mathbf{s}_0\Vert$) in the GVT task when $p^{FA}=.01$.}
    \vspace{-4pt}
    \begin{tabular}{c|c|c}
     & 	$p^{TD}$ ($\chi^2$)  & $p^{TD}$ (learning-based)  
     \\
    \hline
     	  $\Vert \mathbf{s}_0\Vert=.001$    & 	 .009 & 	 .011  \\ \hline 
      	 $\Vert \mathbf{s}_0\Vert=.01$   & 	.009 & 	 .014 \\ \hline 
      	 $\Vert \mathbf{s}_0\Vert=.1$   & 	.26 & 	 .55 \\
    \hline
    \end{tabular}
    \label{tab:GVT_stealthy}
\end{table}

\begin{table}[!t]
    \centering
    \caption{Detection rate of the $\chi^2$ and learning-based detectors for different values of $\alpha$ (deviation) in $m$ when only images were under attack in the VTOL task, with $p^{FA}=.01$.}
    \vspace{-4pt}
    \begin{tabular}{c|c|c|c|c|c}
     & 	$\alpha=.2$  &   $\alpha=.4$ & $\alpha=.6$  &   $\alpha=.8$&   $\alpha=1$
     \\
    \hline
     	 \makecell{ $p^{TD}$ ($\chi^2$)}    & 	 .01 & 	 .79 & 1 & 	 1 &   1 \\ \hline 
      	\makecell{ $p^{TD}$\\ (learning-based) }  & 	.05 & 	 .42 & 1 & 	 1  &   1\\
    \hline
    \end{tabular}
    \label{tab:Tracking_incon}
    \vspace{-8pt}
\end{table}

\subsection{Attacks on Vertical Take-off and Landing}

\begin{table}[!t]
    \centering
    \caption{Averaged true detection rate of $\chi^2$ and learning-based anomaly detectors for different values of initial condition norm ($\Vert \mathbf{s}_0\Vert$) in VTOL task when $p^{FA}=.01$.}
    \vspace{-4pt}
    \begin{tabular}{c|c|c}
     & 	$p^{TD}$ ($\chi^2$)  & $p^{TD}$ (learning-based)  
     \\
    \hline
     	  $\Vert \mathbf{s}_0\Vert=.001$    & 	 .01 & 	 .01  \\ \hline 
      	 $\Vert \mathbf{s}_0\Vert=.01$   & 	.011 & 	 .01 \\ \hline 
      	 $\Vert \mathbf{s}_0\Vert=.1$   & 	.55 & 	 .79 \\
    \hline
    \end{tabular}
    \label{tab:Vtol_stealthy}
\end{table}

\begin{table}[!t]
    \centering
    \caption{Detection rate of $\chi^2$ and learning-based ADs for different values of $\alpha$ (deviation) in meter when only the image is under attack in GVT task when $p^{FA}=.01$.}
    \vspace{-4pt}
 \begin{tabular}{c|c|c|c|c|c}
     & 	$\alpha=.1$  &   $\alpha=.2$ & $\alpha=.3$  &   $\alpha=.4$ &   $\alpha=.5$
     \\
    \hline
     	  $p^{TD}$ ($\chi^2$)    & 	 .17 & 	 .33 & .49 & 	 .65& 	 .81 \\ \hline 
      	 \makecell{ $p^{TD}$\\ (learning-based) }   & 	.17 & 	 .33 & .49 & 	 .64& 	 .67 \\
    \hline
    \end{tabular}
    \label{tab:Vtol_incon}
    \vspace{-8pt}
\end{table}

The VTOL mission was to land the drone on the \emph{mobile} landing mark after it was detected by the vision module. The dashed red line in Fig.~\ref{fig:drone_fig4} shows the trajectory of the target ground vehicle; when the drone was not under attack, the drone was able to follow the vehicle and land on the marker (drone trajectory is shown in orange line).
The blue line presents the drone trajectory when it was under attack. The black and red stars show the same time on both the under attack and non-attacked trajectories; thus, the attacker was able to deviate the drone $1~m$ away from the marker when the ground vehicle got out of range of the camera.

In this experiment, we also assumed that the elements of the initial error vector $\mathbf{s}_0$ were zero except the one associated with the roll angle which was $0.01$. Similar to the GVT,  Fig.~\ref{fig:drone_fig5} and Fig.~\ref{fig:drone_fig6} show the average alarm rate at each time step for 10 experiments; for $\chi^2$ and learning-based anomaly detectors, respectively. Assuming the attack starts at time $t=0$,  the average true detection rate (for $t > 0$) $p^{TD}=0.01$  was almost the same as the average false detection rate (for $t\geq 0$) for both detectors, demonstrating the attack's $\epsilon$-stealthiness.

Moreover, Table~\ref{tab:Vtol_stealthy} captures the impact of the norm of the initial error vector $\mathbf{s}_0$ on the attack stealthiness level --  
the true detection rate 
increases with the increase in the norm of 
$\mathbf{s}_0$. Finally, Table~\ref{tab:Vtol_incon} shows the true detection rate for different deviation levels for the VTOL task. As for the GVT, the attack will be detected if the attacker only compromises the image without consistently attacking the sensors.




%% file: Conclusion.tex
\vspace{-1.2pt}
\section{Conclusion}\label{sec:conclusion}
\vspace{-2pt}

In this work, we have analyzed vulnerability of quadcopters to stealthy  attacks on both camera images and other sensors' measurements. We have provided an attack design framework to consistently falsify both the images and physical sensors data. Specifically, we have shown that the attacker has to exploit knowledge of the system dynamics to derive the suitable 
falsified images and sensing measurements, for the attack to be effective while remaining stealthy from any anomaly detector. 
In the experiments, we have considered two different tasks --  ground vehicle tracking (GVT) and vertical take off and landing (VTOL) and evaluated the impact of the attacks on that system. To detect the presence of attack we have considered a physics-based ($\chi^2$) and a learning-based anomaly detector. We have shown that the attacks result in significant drone trajectory deviations while remaining stealthy from the employed anomaly detectors.